\documentclass{article} 
\usepackage[preprint]{colm2025_conference}
\usepackage{booktabs}
\usepackage{multirow} 

\usepackage{makecell}
\usepackage{soul}
\usepackage{stfloats}
\usepackage{color}
\usepackage{xcolor}
\definecolor{darkgreen}{RGB}{83,129,53}
\definecolor{darkred}{RGB}{163,21,21}

\usepackage{tikz}
\usepackage{subfig}
\usepackage{tcolorbox}
\usepackage{pgfplots}
\usepackage{wrapfig}
\usepackage{amssymb,mathrsfs}
\usepackage{amssymb}
\usepackage{array}
\usepackage{pgfplotstable}
\usepackage{pgf}
\usepackage[normalem]{ulem}
\usepackage{colortbl}
\usepackage{ulem}
\usepackage{xspace}

\usepackage{listings}

\usepackage{scalerel} 
\usepackage{times}
\usepackage{latexsym}
\usepackage{graphicx}
\usepackage{amsmath}
\usepackage{multirow}
\usepackage{amsthm}
\usepackage{amsfonts}
\usepackage{bm}
\usepackage{booktabs}
\usepackage{algorithm}
\usepackage{algpseudocode}
\usepackage{verbatim}
\usepackage{tabularx}
\newcolumntype{L}{>{\raggedright\arraybackslash}X}
\newcommand{\thinkrow}[1]{\rowcolor{yellow!10} \texttt{<think>} #1 \texttt{</think>} \\}
\newcommand{\searchrow}[1]{\rowcolor{green!10} \texttt{<tool\_call>} #1 \texttt{</tool\_call>} \\}
\newcommand{\resultrow}[1]{\rowcolor{white} \texttt{<tool\_response>} #1 \texttt{</tool\_response>} \\}
\newcommand{\answerrow}[1]{\rowcolor{blue!10} \texttt{<answer>} #1 \texttt{</answer>} \\}

\usepackage{tcolorbox}
\tcbuselibrary{skins,breakable,listings}

\tcbuselibrary{breakable} 
\newtcolorbox{examplebox}[2][]{ 
    breakable, 
    colback=white, 
    colframe=cyan, 
    coltitle=white, 
    fonttitle=\bfseries, 
    title=#2, 
    overlay middle={\draw[cyan, line width=1pt](frame.south west)--(frame.south east);}, 
    overlay last={\draw[cyan, line width=1pt](frame.south west)--(frame.south east);}, 
    #1 
}
\usepackage{listings}
\lstdefinelanguage{text}{}
\usepackage{times}
\usepackage{latexsym}

\usepackage[T1]{fontenc}

\usepackage[utf8]{inputenc}

\usepackage{microtype}

\usepackage{inconsolata}

\usepackage{graphicx}

\usepackage{multirow}
\usepackage{booktabs}
\usepackage{adjustbox}

\usepackage{booktabs}   
\usepackage{threeparttable} 
\usepackage{graphicx}   
\usepackage{xcolor}     

\usepackage{amssymb}
\usepackage{microtype}
\usepackage{hyperref}
\usepackage{adjustbox}
\usepackage{color}
\usepackage{xcolor}
\usepackage{tcolorbox}
\usepackage{colortbl}
\usepackage{multicol}
\usepackage{url}
\usepackage{booktabs}
\usepackage{amsmath}
\usepackage{enumitem}
\usepackage{graphicx}
\usepackage{lineno}
\usepackage{xspace}
\usepackage{algorithm}
\usepackage{algpseudocode}
\usepackage{array}
\usepackage{booktabs}
\usepackage{longtable}
\usepackage{arydshln}
\usepackage{caption}

\definecolor{darkblue}{rgb}{0, 0, 0.5}
\hypersetup{colorlinks=true, citecolor=darkblue, linkcolor=darkblue, urlcolor=darkblue}

\title{\centering \raisebox{-.25\height}{\includegraphics[width=0.7cm]{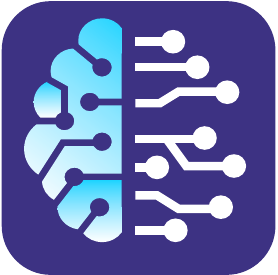}} KnowCoder-A1: Incentivizing Agentic Reasoning Capability with Outcome Supervision for KBQA}

\author{Zhuo Chen$^{1,2,3}$,  Fei Wang$^{1,2}$\thanks{Corresponding authors}\ \ , 
Zixuan Li$^{1,2}$\footnotemark[1]\ \ , Zhao Zhang$^{1,2}$, \\
\textbf{Weiwei Ding$^{1,2}$,}
\textbf{Chuanguang Yang$^{1,2}$,}  
\textbf{Yongjun Xu$^{1,2}$,}
\textbf{Xiaolong Jin$^{1,2,3}$,}
\textbf{Jiafeng Guo$^{1,2,3}$}\\
$^1$Institute of Computing Technology, Chinese Academy of Sciences \\
$^2$State Key Laboratory of AI Safety \\
$^3$School of Computer Science, University of Chinese Academy of Sciences\\
\texttt{\{wangfei@ict.ac.cn, lizixuan@ict.ac.cn\}}
}

\newcommand{\ours}{\textsc{KnowCoder-A1}}
\definecolor{cvprblue}{rgb}{0.21,0.49,0.74}

\begin{document}



\maketitle

\vspace{-10mm}
\begin{center}
    \url{https://ict-goknow.github.io/KnowCoderA1} \\
\end{center}



\begin{abstract}
Knowledge Base Question Answering (KBQA) aims to answer natural-language questions over a structured Knowledge Base (KB). Recent work improves KBQA by adopting an agentic reasoning paradigm, in which Large Language Models (LLMs) iteratively decompose a question, generate its corresponding logical queries, and interact with the KB to derive the answer. However, these methods typically fine-tune LLMs on reasoning trajectories synthesized via process supervision, which offers weak incentives for exploration and thus fails to strengthen the agentic reasoning ability. In this paper, we propose \ours, an LLM that can autonomously perform agentic reasoning on KBs to obtain answers. To incentivize autonomous exploration, \ours\space trains the LLM under outcome-only supervision via a multi-stage curriculum reinforcement learning with an easy-to-hard curriculum. To establish foundational agentic capabilities, \ours\space first fine-tunes the LLM on a small set of high-quality trajectories obtained through outcome-based rejection sampling. Then, to alleviate the reward sparsity inherent in outcome-only supervision, it applies multi-stage curriculum RL with reward schedules that progress from easy to hard. Trained with outcome-only supervision, \ours\space exhibits powerful reasoning behaviors and consistently outperforms prior approaches across three mainstream datasets. Notably, on the zero-shot subset of GrailQA, \ours\space achieves up to an 11.1\% relative improvement while using only one-twelfth of the training data, demonstrating strong agentic reasoning capabilities. Code is available at \url{https://github.com/ICT-GoKnow/KnowCoder-A1}.

\renewcommand{\thefootnote}{}
\renewcommand{\thefootnote}{\arabic{footnote}}%

\end{abstract}

\section{Introduction}
\label{sec:intro}

Knowledge Base Question Answering (KBQA) leverages the rich semantic information in Knowledge Bases (KB) to deeply understand natural language questions and provide precise answers. Since KBQA holds significant value in applications such as search engines, intelligent healthcare, and financial risk control~\citep{jang2017kbqa, wu2024medical, chen2025ifqa}, it has garnered extensive attention in recent years. Although KBQA performs well on simple questions, it still struggles with complex questions in real-world scenarios. This difficulty primarily arises from generalizing to a wide range of unseen complex questions and adapting to the heterogeneous structures of KBs.

\begin{figure*}
    \centering
    \includegraphics[width=1\textwidth]{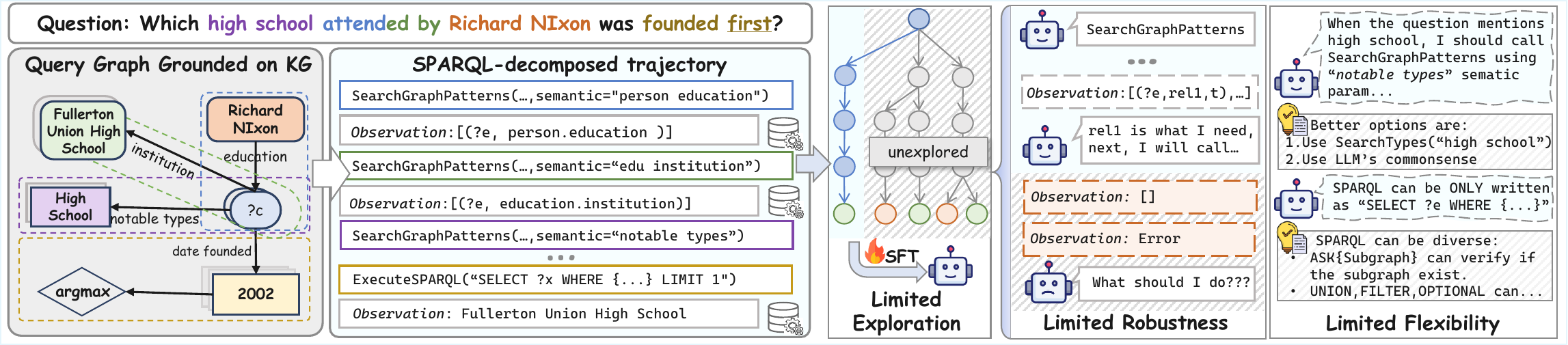}
    \caption{The key limitations of existing agentic approaches that rely on process supervision. }
    \label{fig:intro}
\end{figure*}

Existing semantic-parsing-based~\citep{shu2022tiara, jiangunikgqa} and information-retrieval-based KBQA approaches~\citep{liu2022graftnet, sun2019pullnet} rely on predefined workflows (e.g., generate-then-retrieve or retrieve-then-rank pipelines), resulting in error propagation and adaptability limitations. Recently, a series of approaches has adopted an agentic reasoning paradigm. As illustrated in the middle of Fig.~\ref{fig:intro}, under this paradigm, LLMs autonomously perform task planning, tool calling, and ultimately derive precise answers through multi-turn interactions with KBs. Specifically, existing agentic approaches fall into two categories, i.e., prompting-based~\citep{xiong2024interactive, sunthink} and finetuning-based approaches~\citep{jiang2024kgagent, fang2024dara, zhu2025knowagent}. The former uses hand-crafted few-shot prompts to guide reasoning, relying on the intrinsic reasoning ability of LLMs. However, few examples are inadequate for the model to understand complex questions, perceiving the KB structure, resulting in low performance. In response, finetuning-based approaches adapt open-source LLMs by constructing agentic reasoning trajectories and fine-tuning the model to follow them. However, most of them rely on process supervision. Since manually creating high-quality trajectories is labor-intensive, they typically synthesize reasoning trajectories by decomposing gold logical forms (e.g., SPARQL). Besides, they learn from trajectories via Supervised Fine-Tuning (SFT) to employ the model with agentic reasoning ability. 

Despite the progress, existing finetuning-based approaches rely heavily on process supervision both in data generation and model training. They encourage LLMs to exploit idea trajectories rather than autonomously exploring alternative ones. Consequently, they have two critical limitations: 
\begin{itemize}[leftmargin=*]
\item  \textbf{Limited Robustness.} Under process supervision, as shown in the left part of Fig. \ref{fig:intro}, training trajectories are typically decomposed from gold logical forms, with each step being idealized and error-free. However, such trajectories omit the noise and uncertainty inherent in real-world interactions. As a result, models always struggle when exposed to the noisy, such as tool-call failures and empty results, as they are trained on ideal trajectories. 
\item \textbf{Limited Flexibility.} For many questions, there exists more than one correct trajectory, and in some cases, certain trajectories may even outperform the gold logical forms. Training on trajectories only derived from gold logical forms constrains the model to low-diversity behavioral patterns. As shown in Fig.~\ref{fig:intro}, when the model is presented with an example problem containing a``high school'' type constraint, it insists on finding a predicate with the meaning ``notable type'' and applies the type constraint. However, more efficient solutions exist, such as leveraging common sense to identify entities of a specific type from the candidate set, or using ``SearchTypes'' to more directly constrain the entity types.
\end{itemize}


Motivated by this, in this paper, we propose \ours, an LLM capable of autonomously performing task decomposition, invoking tools, refining formal queries, and executing them to obtain answers. To incentivize autonomous exploration, \ours\space adopts a multi-stage curriculum reinforcement learning approach that relies mainly on outcome supervision. First, to break free from the dependency on process annotation, it employs an outcome-based rejection sampling strategy to curate high-quality trajectories and perform to endow the model with a foundational exploration capability. To mitigate the reward sparsity induced by outcome-only signals, \ours\space employs multi-stage curriculum RL on reward schedules that progress from easy to hard, steadily strengthening autonomous exploration and complex reasoning. Trained with only around 6700 outcome-supervised samples, \ours\space consistently outperforms previous agentic KBQA approaches across three mainstream datasets, i.e., WebQSP, CWQ, and GrailQA. Using \textbf{12×} less training data, it achieves an F1 score of 80.5\% on GrailQA, achieving a \textbf{3.3\%} relative improvement over KBQA-o1, the previous SOTA agentic-based approach. Notably, on the zero-shot subset of GrailQA, \ours\space achieves a relative improvement of up to \textbf{11.1\%} compared to KBQA-o1, demonstrating the strong generalization ability of the proposed approach.

In summary, our contributions are as follows:
\begin{itemize}
    \item  We propose \ours, our first agentic reasoning model for KBQA. Relying mainly on outcome supervision, it learns to act as a robust and flexible agent, capable of recovering from errors and strategically exploring diverse reasoning trajectories.

    \item To address the reward sparsity inherent in outcome-only supervision, we propose a multi-stage curriculum reinforcement learning with an easy-to-hard curriculum, progressing from easy to hard reward criteria while discouraging reward hacking.
    
    \item Extensive experiments on three mainstream datasets demonstrate that \ours\space consistently outperforms previous agentic KBQA approaches, particularly in challenging zero-shot scenarios.
\end{itemize}

\section{Related Work}

\subsection{Traditional KBQA methods} 
Traditional KBQA methods can be categorized into two primary kinds of approaches: Information Retrieval-based (IR-based) and Semantic Parsing-based (SP-based) ones. IR-based methods~\citep{miller2016kvmem, liu2022graftnet, sun2019pullnet, saxena2020embd, shi2021transfernet, zhang2022subgraph} first retrieve query-relevant information from the KB and subsequently rank candidate answers. Among IR-based methods, some~\citep{bordes2015large, chen2019bidirectional} incorporate a long-term memory component that can be read from and written to, or perform a path-walking process over the KB to derive the answer~\citep{qiu2020stepwise, ren2021lego}. SP-based methods~\citep{sun2020sparqa, chen2019uhop, lanquery, jiangunikgqa, ye2022rng, shu2022tiara} transform the natural language question into a logical query, link the entities and relations, and execute the completed query to obtain the answer. With the advent of LLMs, they have been employed to enhance both approaches, serving as powerful subgraph retrievers~\citep{zhao2024kg, liu2024dual} or semantic parsers~\citep{li2023few, nie2024code}. However, these enhancements do not fundamentally alter the predefined, multi-stage workflow, which brings inflexibility and error propagation.
\subsection{Agentic KBQA methods}
Recently, a new paradigm of agentic reasoning has emerged, which views the LLM as an agent that uses tools to interact with the KB, progressively reason, and finally find the answer. We call them agentic methods in this paper. They can be divided into two main categories: prompt-based and finetuning-based ones. The former methods, such as Interactive-KBQA~\citep{xiong2024interactive}, QueryAgent~\citep{huang2024queryagent}, Readi~\citep{cheng2024call}, and TOG~\citep{sunthink}, leverage carefully designed interaction strategies and prompts to elicit the model's reasoning capabilities. However, these methods struggle to equip the model with sufficient agentic capability to effectively perceive the KB and execute complex reasoning. Their success is also heavily dependent on the intrinsic capabilities of LLMs, most of which are closed-source and have an extremely large number of parameters. However, the limited number of examples is insufficient for the model to fully comprehend complex questions and capture the underlying KB structure, leading to suboptimal performance. Finetuning-based methods aim to distill reasoning abilities into smaller, open-source LLMs. Approaches~\citep{gu2023don,jiang2024kgagent,zhu2025knowagent, fang2024dara} achieve this by constructing high-quality reasoning trajectories and finetuning on them. However, these trajectories, by decomposing and augmenting gold program annotations and the reliance on process supervision via SFT, limit the model's capacity for autonomous exploration, resulting in poor robustness and flexibility. Although some recent works have used Monte Carlo Tree Search (MCTS) methods~\citep{luokbqao1, xiong2025mcts} to expand samples during inference for incremental fine-tuning, this is an indirect way to foster exploration and still operates within the confines of the process-supervised SFT paradigm. Overall, existing finetuning-based methods fall under SFT-based process supervision, which, as noted in Section \ref{sec:intro}, struggles to encourage autonomous exploration, leading to less robust and adaptable reasoning trajectories.
\section{Preliminaries}\label{sec:prelim}

\textbf{Knowledge Base.}
A Knowledge Base (KB) is a structured knowledge graph $\mathcal{G}=(\mathcal{E},\mathcal{R},\mathcal{F})$, where $\mathcal{E}$ is the set of entities, $\mathcal{R}$ is the set of relations, and $\mathcal{F}$ is the set of factual triples. Each triple $f \in \mathcal{F}$ takes the form $(h,r,t)$, where $h, t \in \mathcal{E}$ are the head and tail entities, and $r \in \mathcal{R}$ is the relation.

\textbf{KBQA.}
Given a natural-language question $q$, the KB $\mathcal{G}$, and a set of topic entities $E_q\subseteq\mathcal{E}$ mentioned in $q$, the objective of the KBQA task is to find the answer set that maximizes the conditional probability $P(\mathcal{A}_q | q, E_q, \mathcal{G})$. 
Notably, following prior work~\cite{luokbqao1}, we assume entity mentions in $q$ are linked to the knowledge graph $\mathcal{G}$ and given as an input.

\textbf{Agentic Reasoning Paradigm for the KBQA Task.}
In the agentic KBQA paradigm, the LLM is viewed as an agent learning a policy $\pi$ to interact with the environment. At each step $t$, the agent chooses an action $a_t \in \textsc{Action}$ based on the question $q$ and its historical reasoning trajectory $\tau_t = \{(c_1,a_1,o_1),\ldots,(c_{t},a_{t},o_{t})\}$, where $\textsc{Action}$ denotes the action space, $c_t$ denotes the explicit thinking process before the action, and $o_t$ is the observation from the environment. This interactive process continues until the agent finds the final answer set $\mathcal{A}_q$ or reaches the maximum number of steps.

The objective is to learn an optimal policy $\pi^*$ that maximizes the probability of generating a trajectory $\tau$ that leads to the gold answer set $\mathcal{A}^*_q$. This can be expressed as maximizing the total probability of all successful trajectories:
\begin{equation}
\pi^* = \arg\max_{\pi} \sum_{\tau \in \mathcal{T}_{success}} P(\tau | q, E_q, \mathcal{G}; \pi),
\end{equation}
where $\mathcal{T}_{success}$ is the set of all trajectories that successfully yield the gold answers $\mathcal{A}^*_q$, and the trajectory probability $P(\tau | \dots; \pi)$ is determined by the policy $\pi$.
\section{Methodology}\label{sec:method}

\begin{figure*}[!t]
    \centering
    \includegraphics[width=\textwidth]{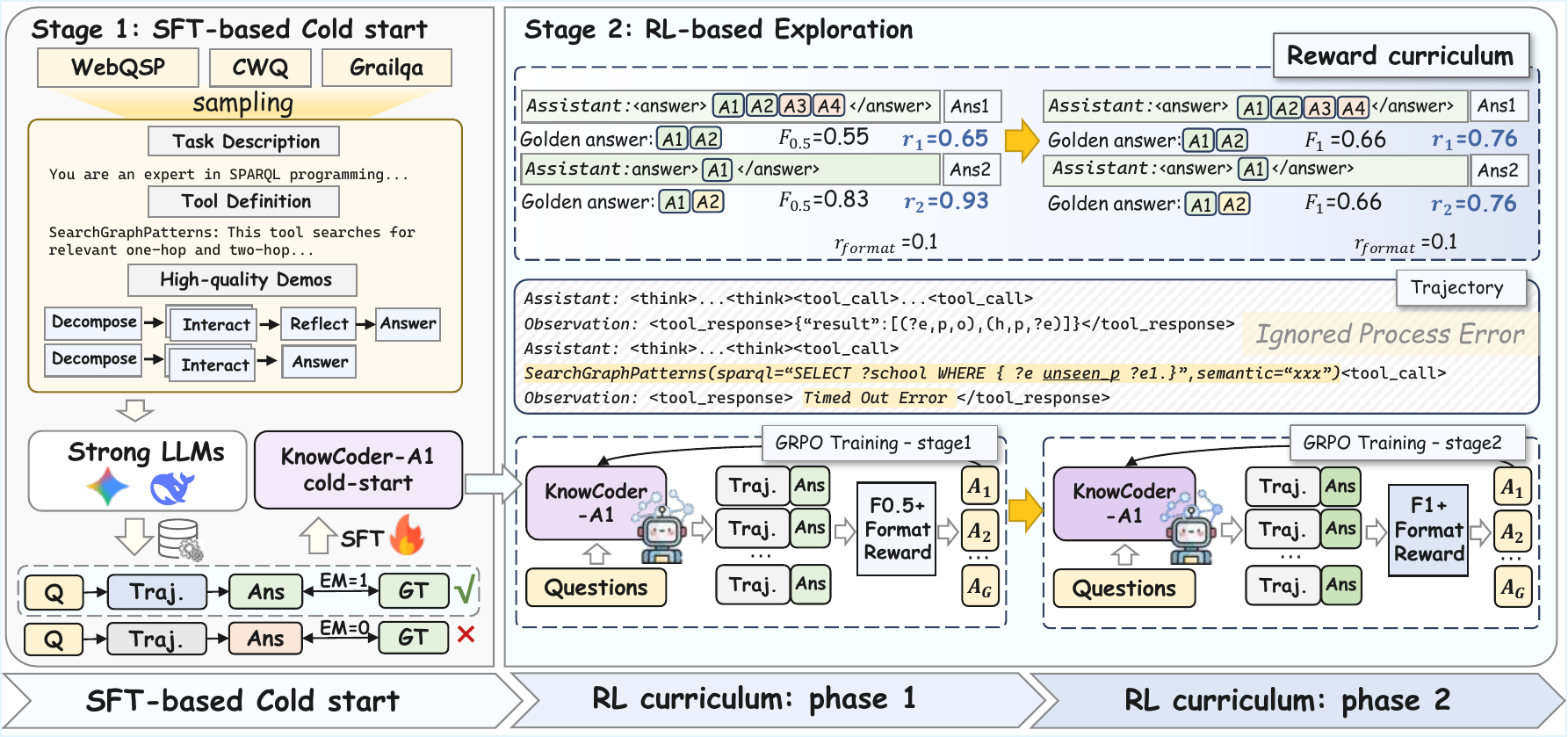}
    \caption{An overview of the training framework of \ours \space. Stage 1 (left): the SFT-based cold-start process, where high-quality trajectories are curated from strong LLMs to fine-tune an initial agent. Stage 2 (right): the multi-phase Reinforcement Learning curriculum, where the agent is progressively improved through exploration and a dynamic reward strategy.}
    \label{fig:main}
\end{figure*}

\subsection{Agentic Initialization}
\ours \space follows a ReAct-style~\citep{yao2023react} agent paradigm that interleaves explicit reasoning and tool use over a KB environment, and terminates with a final answer.

\textbf{Environment.} We view the KB and the tool executor as the environment:
$\mathcal{M}=(\mathcal{G},\ \texttt{Exec})$,
where \texttt{Exec} executes tool calls against $\mathcal{G}$ and returns structured observations (including error diagnostics when execution fails). It is worth noting that error messages are retained to provide corrective signals to help LLM refine subsequent tool calls.

\textbf{Action Space.} At each step, the agent first produces an explicit thinking process and then makes a tool call from a finite action set $\textsc{Action}$. We instantiate three tools based on Interactive-KBQA~\citep{xiong2024interactive}:
\begin{itemize}
    \item \textsc{SearchTypes}(typename): Given a type name, retrieves and ranks candidate types in similarity order.
    \item \textsc{SearchGraphPatterns}(sparql, semantic): Given a partial SPARQL sketch (assumed to start with \texttt{SELECT DISTINCT ?x WHERE \{...\}}), returns up to 10 one- and two-hop subgraphs in which the variable \texttt{?x} appears as the head or tail object. An optional \texttt{semantic} parameter can be provided to rank the returned subgraphs in similarity order.
    \item \textsc{ExecuteSPARQL}(sparql): Given a SPARQL query, executes it against a sandboxed endpoint over $\mathcal{G}$ and returns the result. This enables flexible query on the graph, such as hypothesis testing (e.g., subgraph existence) or end-to-end query execution.
\end{itemize}

\textbf{State and Trajectory.}
Let $c_t$ denote the explicit thinking process in natural language, $a_t\in \textsc{Action}$ the action, and $o_t$ the observation at step $t$. The trajectory, which also serves as the agent's state, is the history of all past interactions. The trajectory at step $t$ is denoted as:
\begin{equation}
\tau_t = \{(c_1,a_1,o_1),\ldots,(c_{t},a_{t},o_{t})\}.
\end{equation}

\textbf{Interactive Process.}
Given a question $q$, the agent begins with an initial prompt 
\begin{equation}
\mathrm{Prompt}=\big(\mathrm{Inst},\ q,\ E_q\big),
\end{equation}
where the instruction consists of a brief task description, tool definitions, and the required interaction format, followed by the question $q$ and its topic entities $E_q$. The specific prompt is provided in the Appendix~\ref{app:prompt}.

At each step $t$, the agent first generates a thought $c_{t+1}$ and a tool call $a_{t+1}$ conditioned on the current trajectory and the initial prompt:
\begin{equation}
c_{t+1}, a_{t+1} = \mathrm{LLM}(\tau_t, \mathrm{Prompt}).
\end{equation}

The tool call $a_{t+1}$ is then passed to the executor, which returns an observation $o_{t+1} = \texttt{Exec}(a_{t+1})$. The new interaction turn $(c_{t+1}, a_{t+1}, o_{t+1})$ is appended to the current trajectory to form the trajectory for the next step, $\tau_{t+1}$. This iterative process continues until the agent determines it has gathered sufficient information and decides to output the final answer, or reaches $T$, the maximum number of steps.

\subsection{SFT-based Cold-start Stage}
Due to the heterogeneous structures of KBs and the complexity of questions, employing a base model with insufficient reasoning and tool-use capabilities as an agent makes it difficult to guarantee the discovery of trajectories to the correct answer. To mitigate this issue, we introduce a cold-start stage where the agent is fine-tuned on reasoning trajectories to acquire foundational perception and reasoning capabilities. Instead of relying on program-decomposed trajectories, we generate trajectories via outcome-based rejection sampling. Moreover, to avoid overfitting while ensuring sufficient guidance, we adhere to the principle of constructing a dataset that is concise, yet high in quality and diversity. In the following, we will first describe this data curation process and then detail the model's training procedure. 

\subsubsection{Data Construction.} 
As discussed in Sec.~\ref{sec:intro}, program-decomposed trajectories used in existing approaches fundamentally lack exploration, resulting in the limited robustness and flexibility of the model. 
To overcome the limitations and obtain a high-quality and diverse dataset, we design an outcome-based data construction strategy, the overall pipeline is illustrated in the left panel of Fig.~\ref{fig:main}, which contains three core components:

First, to ensure the model learns a diverse set of fundamental reasoning patterns, we curate a mixed question set using a principled sampling strategy across several sources. From WebQSP, we balance one- and two-hop questions to teach varied reasoning depths. From CWQ, we up-sample infrequent question types to guarantee exposure to complex compositional structures. Finally, from GrailQA, we up-sample questions without topic entities to improve the model's ability to generalize.

Next, to endow the model with stronger exploratory abilities and expand its exploration space, we use a prompt designed to encourage trial-and-error. As the yellow block in Fig.~\ref{fig:main} shows, the prompt contains three parts: task description, tool definition, and high-quality demos: a few manually crafted exemplars that demonstrate a general agentic reasoning process, including planning, interacting, and self-correction after errors. Then, we prompt strong models (e.g., Gemini 2.5 Flash  ~\citep{comanici2025gemini} and DeepSeek-V3  ~\citep{guo2025deepseek}) to generate multiple candidate trajectories for each question.

After that, to ensure the correctness of the trajectories, we apply an outcome-based rejection sampling, filtering the generated trajectories using two criteria, retaining only those that are both correct and fully grounded in evidence:
\begin{itemize}
\item \textbf{Exact-match Correctness.} The Exact Match (EM) score~\citep{talmor2018web} between the predicted and gold answers must be $1$. This ensures that at least one of the predicted answers appears in the gold answers, which largely guarantees the correctness of the trajectory.
\item \textbf{Evidence Grounding.} All predicted answers must appear in the observation history $\{o_0,\ldots,o_T\}$, ensuring that the agent does not rely on internal parametric knowledge that could lead to hallucinations during training.
\end{itemize}

Finally, we apply a final filtering step to these trajectories. We ensure that each question has a maximum of three trajectories and that the number of trajectories is equal or balance across the aforementioned question types. The statistics of the resulting cold-start dataset can be found in Table~\ref{tab:training_samples} in Appendix~\ref{app:stat}, where the size of the dataset is less than most of the existing methods.

\subsubsection{Multi-Turn Finetuning}
In the fine-tuning stage, we aim to teach the agent to reason over long, complex interaction histories. To achieve this efficiently without losing critical long-term context, we fine-tune the agent on complete trajectories. This approach avoids the disadvantage of alternative methods, which split the trajectory into multiple input-output steps: retaining the full context for each step is computationally inefficient, while using only a limited history window risks severing crucial long-term dependencies.

Specifically, we use the entire history as input but compute the loss only on the agent's generated tokens, masking out the observation part returned by tools during training. This prevents these external tokens from affecting the loss calculation, ensuring that the retrieved results do not interfere with the model's internal reasoning and generation processes. The process can be formalized as follows: 


We define the input prompt as $x$ and the expert trajectory as $\tau$. Here the trajectory $\tau$ is represented as a sequence of $M$ tokens, $\tau = (y_1, \dots, y_M)$, which comprise the expert's thoughts and actions. The agent is fine-tuned by maximizing the likelihood of generating this expert trajectory conditioned on the input $x$. During training, we specifically mask out observation tokens, computing the loss only on the expert's thought and action tokens. This objective is formulated as:
\begin{equation}
\mathcal{L}_\tau(\theta) = -\sum_{j=1}^{M} \log P_\theta(y_j | x, y_{<j}),
\end{equation}
where $\theta$ represents the model's parameters, $M$ is the total number of tokens in the sequence, $y_j$ is the $j$-th token, and $y_{<j}$ denotes the sequence of preceding tokens $(y_1, \dots, y_{j-1})$.

\subsection{RL-based Exploration Stage}
Based on the cold-start model, in the RL stage, we aim to further promote autonomous exploration and eliminate the reliance on process supervision. To achieve this, we employ the outcome-based Group Relative Policy Optimization (GRPO) method~\citep{shao2024deepseekmath}, where any trajectory leading to a correct answer receives a positive incentive, thereby encouraging the agent to explore and acquire stronger reasoning capabilities. Moreover, to ensure the RL training is both balanced and effective, we implement two key strategies. 

First, to balance the agent's exploration across question types (e.g., simple vs. complex), we sample a balanced set of questions for the RL training, following a similar principle as in the cold-start stage. Second, to mitigate the reward sparsity induced by outcome-only signals and refine the training process, we introduce a curriculum strategy on the reward, scheduling the reward strictness, guiding the learning process from easy to hard and effectively suppressing reward hacking.

In this section, we first introduce the GRPO method, followed by the design of our data selection and reward curriculum strategies.

\subsubsection{Policy Optimization via GRPO}
Inspired by the recent success of policy gradient methods on reasoning tasks~\citep{jin2025search,xue2025simpletir,luo2025graph}, we employ the GRPO method for optimization. GRPO estimates the baseline using the rewards from a group of rollouts for the same query to compute the advantage. Its objective function can be expressed as follows:

\begin{align}
\mathcal{J}_{\mathrm{GRPO}}(\theta)
&=
\mathbb{E}_{\substack{q \sim \mathcal{D}_Q\\ \{\tau^{(i)}\}_{i=1}^N \sim \pi_{\theta_{\text{old}}}(\cdot|q)}}
\Bigg[
  \frac{1}{N}\sum_{i=1}^N \frac{1}{|\tau^{(i)}|}
  \sum_{j=1}^{|\tau^{(i)}|}
  \min(\rho_{i,j}(\theta)\hat{A}_{i,j}, \operatorname{clip}(\rho_{i,j}(\theta), 1 \pm \epsilon)\hat{A}_{i,j})
\Bigg]
\\[10pt]  
&\nonumber \qquad -\,\beta\,\mathbb{D}_{\mathrm{KL}}(\pi_\theta \,\|\, \pi_{\text{ref}}),
\end{align}
\begin{align}
\hat{A}_{i,j} &=  
\frac{R(\tau^{(i)}) - \operatorname{mean}\big(\{R(\tau^{(i)})\}_{i=1}^N\big)}
     {std\big(\{R(\tau^{(i)})\}_{i=1}^N\big)},
\end{align}
\begin{align}
\rho_{i,j}(\theta) = \frac{\pi_\theta(\tau_{j}^{(i)} | \tau_{<j}^{(i)})}{\pi_{\theta_\text{old}}(\tau_{j}^{(i)} | \tau_{<j}^{(i)})}.
\end{align}

Here, $q$ is the question from the train set $D_Q$ , and $\tau^{(i)}$ is the $i$-th trajectory generated by the policy model $\pi_{\theta_{old}} $for $q$. $R(\tau^{(i)})$ is a scalar reward derived solely from the final outcome of the trajectory. $\hat{A}_{i,j}$ represents the group-wise relative advantage, which is estimated by normalizing the group-level rewards for the same query, normalized by the standard deviation $std(\cdot)$ to stabilize training. $\rho_{i,j}(\theta)$ denotes the token-level importance ratio. $\pi_{\text{ref}}$ is a reference policy (i.e., the model after the cold-start stage), and $\beta$ controls the strength of the KL loss $\mathbb{D}_{\mathrm{KL}}(\cdot || \cdot)$ towards the reference policy.

In this paper, we adopt an on-policy variant of GRPO. This variant applies a single SGD step to the entire rollout batch, which sets the importance ratio in the aforementioned formula to 1. This approach has also been shown to be an effective way to incentivize exploration~\citep{he2025skywork}. 

\subsubsection{Reward and Curriculum strategy}
During the earlier phases of RL training, the cold-start model often lack the capability to generate complete and correct reasoning trajectories for complex questions. This results in reward sparsity when relying solely on final outcomes, and meanwhile hinders effective exploration. Thus, we introduce a composite reward function that mitigates reward sparsity by providing denser feedback signals, while guiding the exploration process through a carefully designed easy-to-hard curriculum. The reward is specifically composed of a Format Reward and a multi-phase Answer Reward.

\textbf{Format Reward} ($R_{\text{fmt}}$). To discourage invalid trajectories and ensure that answers can be reliably parsed for reward calculation, we introduce a format reward. This component checks whether a trajectory adheres to the required format, specifically the presence of the answer box \verb|\boxed{}|. If the format is correct, we set $R_{\text{fmt}}=0.1$; otherwise, $R_{\text{fmt}}=0.0$. 

\textbf{Answer Reward} ($R_{\text{ans}}$). To verify the correctness of the predicted answer and avoid a sparse reward signal, we evaluate the agreement between the predictions inside \verb|\boxed{}| and the gold answers using $F_\beta$ score. Compared with a simple EM score, the $F_\beta$ score balances precision and recall via an adoptable hyperparameter $\beta$, which prevents agents from trivially obtaining a high score by returning a large, low-precision candidate set. Concretely, the $F_\beta$ score is computed as:\begin{equation}Precision=\frac{|\widehat{\mathcal{A}}\cap \mathcal{A}_q^\star|}{|\widehat{\mathcal{A}}|},\quad Recall =\frac{|\widehat{\mathcal{A}}\cap \mathcal{A}_q^\star|}{|\mathcal{A}_q^\star|},\quad F_{\beta}=\frac{(1+\beta^2) \cdot Precision \cdot Recall }{\beta^2  \cdot Precision + Recall },\end{equation}
where $\widehat{\mathcal{A}}$ denotes the de-duplicated set of answers produced by the model and $\mathcal{A}_q^\star$ denotes the set of gold answers, and $|\cdot|$ denotes multiset cardinality.

Based on the answer reward, we employ a curriculum learning strategy by adapting the $\beta$ parameter. This gradually adjusts the reward difficulty from easy to hard, as illustrated in the top-right part of Figure \ref{fig:main}.
\begin{itemize}
    \item \textbf{Phase 1: Precision-focused Reward.} In early training, we set $\beta=0.5$ in the answer reward. This precision-leaning objective grants relatively high reward for correct items even when recall is incomplete. As the Fig.~\ref{fig:main} shows, if the golden answer is \{A1, A2\}, a precise but partial prediction \{A1\} achieves a high reward ($r_2=0.83$). Conversely, an prediction that includes extraneous items, such as \{A1, A2, A3, A4\}, is penalized with a lower reward ($r_1=0.55$), encouraging accurate predictions and preventing the model from hacking the reward by emitting large candidate sets.
    \item \textbf{Phase 2: Balanced Reward.} Once the model achieves stable precision, we switch the objective to the $F_1$ score ($\beta=1$), which balances precision and recall. Under this balanced metric, a partial prediction is less rewarded than in Phase 1, incentivizing the agent to explore more thoroughly and recover the full set of correct answers.
\end{itemize}
The total reward is the sum of the two parts, capped at a maximum value of 1.0:
\begin{equation}
R_{\text{total}} \;=\; min(R_{\text{fmt}} \;+\; R_{\text{ans}}, 1.0).    
\end{equation}

\section{Experiments}\label{sec:experiment}

To prove the effectiveness of \ours, we conduct comprehensive experiments and present the setup, main results, and analysis in this section. Specifically, we focus on the following research questions (RQ): \textbf{RQ1:}
Does \ours \space outperform other directly comparable methods? \textbf{RQ2:} Does the main components of \ours \space work effectively? \textbf{RQ3:} How does \ours \space progressively enhance the agent's reasoning capability for the KBQA task with outcome-only supervision? \textbf{RQ4:} Does \ours \space generates flexible and robust reasoning trajectories?

\subsection{Experiment Setup}
\textbf{Dataset and Metrics}. We evaluate \ours \space on three widely-used KBQA datasets: WebQSP, CWQ, and the generalization-focused GrailQA, using a subset of the official testing data. Following existing work~\citep{xiong2024interactive}, we primarily evaluate the F1 metrics, supplemented by Random Hits@1(RHits@1) and EM score. As our approach does not rely on semantic parsing, we do not report logical form accuracy~\citep{gu2021beyond}. 

\textbf{Baselines}. We compare \ours \space against two main categories of methods: prompting-based and fine-tuning-based methods. For the former, we include a diverse range of prompting strategies, from direct prompting, such as IO, CoT, etc., and agent-based reasoning method Interactive-KBQA~\citep{xiong2024interactive}. To ensure a fair comparison and reproducibility, we report the scores for these methods directly from their original publications. For the fine-tuning-based methods, we select SOTA methods finetuned on full data, include RnG-KBQA~\citep{ye2022rng} and TIARA~\citep{shu2022tiara}, and the directly comparable low-source methods, which are finetuned on low-source data, includes agentic approaches like SFT-traj.~\citep{xiong2024interactive}, KBQA-o1~\citep{luokbqao1}, and MCTS-KBQA~\citep{xiong2025mcts}.

Further details on the datasets, baselines, evaluation protocol, and implementation are provided in Appendix~\ref{app:setup}.

\begin{table*}[!t]
  \centering
  \caption{Experiment Results (in percentage) on WebQSP, CWQ, and GrailQA. The \textbf{Bold} and \underline{underlined} numbers indicate the best and second-best low-resource performance.}
  \label{tab:all-three}
  \resizebox{\linewidth}{!}{%
  \begin{tabular}{c c c c c c c c c c}
    \toprule
    \multirow{2}{*}{Method} & \multirow{2}{*}{Backbone} &
      \multicolumn{2}{c}{WebQSP} & \multicolumn{2}{c}{CWQ} & \multicolumn{4}{c}{GrailQA} \\
    \cmidrule(lr){3-4}\cmidrule(lr){5-6}\cmidrule(lr){7-10}
     & & Overall~F1 & RHits@1 & Overall~F1 & EM & I.I.D. & Compositional & Zero-shot & Overall \\
    \midrule
    \multicolumn{10}{c}{\textit{Prompting Methods}} \\
    \midrule
    IO        & GPT-4-turbo     & 39.3 & 45.5 & 33.9 & 45.7 & --- & --- & --- & 29.4 \\
    CoT       & GPT-4-turbo     & 39.7 & 47.5 & 33.7 & 43.7 & --- & --- & --- & 28.1 \\
    CoT+SC    & GPT-4-turbo     & 39.0 & 47.1 & 36.6 & 47.5 & --- & --- & --- & 29.6 \\
    KB-BINDER & Codex-davinci-002 & 52.6 & --- & --- & --- & 43.3 & 36.6 & 44.0 & 42.2 \\
    KB-Coder  & GPT-3.5-turbo     & 55.7 & --- & --- & --- & 45.5 & 38.6 & 47.3 & 44.9 \\
    ARG-KBQA  & GPT-3.5-turbo     & 58.8 & --- & --- & --- & 51.5 & 41.8 & 52.1 & 48.5 \\
    Interactive-KBQA & GPT-4-turbo     & 71.2 & 72.5 & 49.1 & 59.2 & --- & --- & --- & --- \\
    \midrule
    \multicolumn{10}{c}{\textit{Fine-tune-based Methods}} \\
    \midrule
    RnG-KBQA & BERT-base-uncased         & 75.6 & ---   & ---   & ---   & 89.0 & 68.9 & 74.7 & 76.9 \\
    TIARA    & BERT-base-uncased         & 78.9 & 75.2 & ---   & ---   & 91.2 & 74.8 & 80.7 & 81.9 \\
    \midrule
    SFT-traj. & Llama2-7B     & 57.5 & 59.9 & 42.2 & 48.8 & 54.2 & 45.8 & 44.2 & 48.5 \\
    MCTS-KBQA         & Llama-3.1-8B & \underline{76.0} & 76.2 & \underline{66.8} & \underline{75.2} & --- & --- & --- & --- \\
    KBQA-o1           & Qwen2.5-7B   & 57.8 & ---   & ---   & ---   & \textbf{84.4} & \textbf{77.0} & \underline{75.7} & \underline{77.9} \\
    \midrule
    \ours-cold start & Qwen2.5-Coder-7B & 56.1 & 66.3 & 54.8 & 59.8 & 60.6 & 63.5 & 67.2 & 63.1 \\
    \rowcolor{blue!8}
    \ours         & Qwen2.5-Coder-7B & \textbf{77.2} & \textbf{80.1} & \textbf{68.3} & \textbf{75.7} & \underline{81.1} & \underline{70.6} & \textbf{84.1} & \textbf{80.5} \\
    \bottomrule
  \end{tabular}}
\end{table*}

\begin{table}[h]
  \centering
  \small
  \caption{Training instances (Train) and evaluation result with consumption (Evaluation), where 'Time' represents the average number of rollouts required per question at inference, normalized to our method (1x).}
  \label{tab:sample_used}
  \resizebox{\linewidth}{!}{%
  \begin{tabular}{l c c c c c c c c c c}
    \toprule
    \multirow{3}{*}{Model}
      & \multicolumn{4}{c}{Train}
      & \multicolumn{6}{c}{Evaluation} \\
    \cmidrule(lr){2-5}\cmidrule(lr){6-11}
      & \makecell{P.S.\\\#Inst.}
      & \multicolumn{3}{c}{\makecell{O.S.\#Inst.}}
      & \multicolumn{2}{c}{CWQ}
      & \multicolumn{2}{c}{WebQSP}
      & \multicolumn{2}{c}{GrailQA} \\
    \cmidrule(lr){3-5}\cmidrule(lr){6-7}\cmidrule(lr){8-9}\cmidrule(lr){10-11}
      &  & WebQSP & CWQ & GrailQA
      & Time & F1 & Time & F1 & Time & F1 \\
    \midrule
    MCTS\text{-}KBQA
      & 750 & 2743 & \textbf{2843} & \textemdash
      & \(3.24\times\) & 66.8 & \(2.91\times\) & 76.0 & \textemdash & \textemdash \\
    KBQA\text{-}o1
      & 240 & 2929 & \textemdash & 43851
      & \textemdash & \textemdash & \(6\times\) & 57.8 & \(6\times\) & 77.9 \\
    \textbf{Ours}
      & \textbf{0} & \textbf{2414} & \underline{4699} & \textbf{3563}
      & \(1\times\) & \textbf{68.3} & \(1\times\) & \textbf{77.2} & \(1\times\) & \textbf{80.5} \\
    \bottomrule
  \end{tabular}%
  }
\end{table}
\subsection{Main Results - RQ1}
To evaluate the performance of \ours, we conducted comprehensive experiments on three datasets, with the results presented in Table~\ref{tab:all-three}. For additional context on the efficiency of our method, we compare the number of samples used during training with the number of rollouts used during inference in Table~\ref{tab:sample_used}. Taken together, these results demonstrate that \ours establishes a new state-of-the-art for low-resource KBQA across all benchmarks. This shows the effectiveness of our approach, which can be analyzed from three key aspects.

First, \ours \space significantly outperforms all directly comparable low-resource baselines, particularly on the more complex CWQ and the generalization-focused GrailQA datasets. Meanwhile, it nearly closes the gap with fully-supervised methods while using only a fraction of the data (~10k outcome-supervised samples vs. ~60k fully-annotated queries), highlighting its exceptional data efficiency.

Second, results prove that the outcome-supervised multi-stage training strategy is effective than the process-supervised SFT strategy.
Results in the last two blocks of Table~\ref{tab:all-three} indicate that \ours \space our approach consistently outperforms leading process-supervised SFT baselines. On GrailQA, \ours \space achieves an F1 score of 80.5\%, which represents a 3.3\% relative improvement over the prior state-of-the-art method KBQA-o1 while using 12× less training data (3.5k vs. 43.8k samples). While performance is slightly lower on GrailQA's I.I.D. and compositional subsets, which favor extensive training data to cover specific patterns, our method demonstrates a relative improvement of up to 11.1\% on the zero-shot subset. This proves its superior ability to reason effectively on truly unseen questions.

Furthermore, beyond higher overall scores, \ours \space is more efficient in both supervision and inference. In terms of supervision, as detailed in the ``Train'' columns of Table~\ref{tab:sample_used}, \ours \space uses zero process-supervised (P.S.) instances and a comparable or smaller number of outcome-supervised (O.S.) samples than MCTS-based competitors. As montioned before, \ours \space outperforms KBQA-o1 on GrailQA while using much fewer supervised samples. In terms of inference, as shown in the ``Evaluation'' columns of Table~\ref{tab:sample_used}, \ours \space employs a single linear reasoning pass and avoids the costly sampling required by MCTS, reducing latency by 3.2–6 times.
These results support outcome supervision as a more effective and efficient alternative to incremental SFT on pre-defined or explored paths. 

Finally, \ours \space unlocks SOTA reasoning capabilities in smaller models. As shown in the first block of Table~\ref{tab:all-three}, \ours \space consistently outperforms prompting-based methods that rely on significantly larger models like GPT-4-turbo by at least 6\% at F1. This demonstrates an essential improvement in the model's intrinsic reasoning ability.

\begin{table*}[!t]
  \centering
  \caption{Ablation Study on Three KBQA Datasets.}
  \label{tab:ablation}
  \resizebox{\textwidth}{!}{ %
  \begin{tabular}{lcccccccc}
    \toprule
    \multirow{2}{*}{\textbf{Model}} &
      \multicolumn{4}{c}{\textbf{GrailQA}} &
      \multicolumn{2}{c}{\textbf{WebQSP}} &
      \multicolumn{2}{c}{\textbf{CWQ}} \\
    \cmidrule(lr){2-5} \cmidrule(lr){6-7} \cmidrule(l){8-9}
    & IID & Comp. & Zero-shot & Overall F1
    & F1 & Hits@1
    & F1 & EM \\
    \midrule
    \ours              & \textbf{81.1} & 70.6 & \textbf{84.1} & \textbf{80.5} & \textbf{77.2} & 80.1 & \textbf{68.3} & \textbf{75.7} \\
    \midrule
    \multicolumn{9}{l}{w/o reward curriculum} \\
    \quad EM as Reward      & 45.5 & 50.6 & 73.6 & 62.8 & 73.8 & 76.1 & 50.2 & 55.0 \\
    \quad F1 as Reward             & 68.7 & \textbf{77.6} & 78.6 & 76.0 & 71.9 & 77.0 & 58.3 & 59.8 \\
    \quad F0.5 as Reward           & 79.1 & 68.2 & 82.6 & 78.7 & 76.3 & \textbf{80.9} & 67.2 & 74.7 \\
    \quad F1+F0.5 as Reward  & 74.9 & 65.5 & 82.6 & 76.1 & 73.8 & 77.0 & 64.6 & 66.1 \\
    \midrule
    \addlinespace[1pt]
    w/o RL (cold-start)        & 60.6 & 63.5 & 67.2 & 63.1 & 56.7 & 64.9 & 50.5 & 59.8 \\
    w/o upsampling             & 52.5 & 53.0 & 62.3 & 60.2 & 56.1 & 66.3 & 54.8 & 52.7 \\
    \bottomrule
  \end{tabular}
  }
\end{table*}
\subsection{Ablation Study - RQ2}
To verify the effectiveness of our key designs, we conduct a comprehensive ablation study, and the results are presented in Table~\ref{tab:ablation}. First, we assess the reward curriculum by replacing our proposed easy-to-hard ($F_{0.5}$-then-$F_{1}$) strategy, give the result using EM, $F_1$, $F_{0.5}$, and revered strategy $F_1$-then-$F_{0.5}$ as reward, respectively. Next, we evaluate the contribution of the entire RL stage by removing it altogether. Finally, we remove the data sampling strategy from the initial cold-start phase. The results confirm that each component is integral to the model's performance, as removing or modifying any of them leads to a noticeable degradation.

\textbf{Effectiveness of the Easy-to-Hard Reward Curriculum.}
The first block of Table~\ref{tab:ablation} shows the effectiveness of the Easy-to-Hard Reward Curriculum. Take the GrailQA dataset as an example, we observe that an overly simplistic signal like EM is insufficient for exploration; it induces "over-recall reward hacking," resulting in a massive drop in the $F_1$ score on GrailQA. Meanwhile, a stricter F1 reward is too punishing early on, leading to a -4.5\% F1 drop. Notably, reversing the curriculum to a hard-to-easy schedule ($F_1$-then-$F_{0.5}$) also degrades performance (-4.4\% F1 drop). This confirms our hypothesis: starting with an ``easy task'' encourages broad exploration to discover correct reasoning paths, while the subsequent ``hard task'' refines the agent's ability to achieve a better balance between precision and recall.

\textbf{Critical Contribution of the RL Stage.} From the second block of Table~\ref{tab:ablation}, we can observe that removing the entire RL-stage causes at least 17.4\% absolute drop in the F1 score. This large gap suggests that RL-based exploration is essential for refining the agent's reasoning capability beyond initial supervised learning.

\textbf{Importance of Upsampling in the Cold-Start Stage.}
Finally, as the last row shows, removing the up-sampling of infrequent question types leads to poor performance on those specific categories. This underscores the importance of our sampling strategy for building a model that can generalize across a diverse range of reasoning patterns.

\subsection{Analysis - RQ3}

To answer the RQ3 and investigate how \ours \space progressively enhances the agent's reasoning capability under the outcome-only supervision, we analyze from the train dynamics, the evolution of its reasoning robustness and diversity, and the comparative performance against process-supervised methods. 

\subsubsection{Analysis of Training Dynamics with Outcome-Only Supervision}
\label{sec:trian_dym}
\begin{figure*}
    \centering
    \includegraphics[width=1\textwidth]{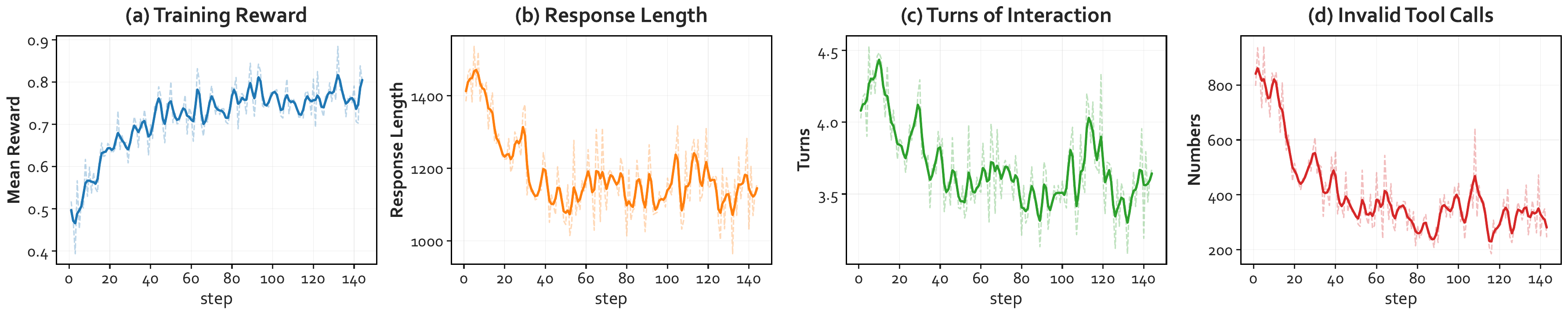}
    \caption{Training curves for \ours, illustrating: (a) training reward, (b) response length, (c) interaction turns, and (d) the number of invalid tool calls.}
    \label{fig:train_dynamic}
\end{figure*}

As shown in Fig.~\ref{fig:train_dynamic}, to track the evolution of the model's capabilities during training, we present the curves of (a) training reward, (b) response length (token), (c) turns of interaction, and (d) the number of invalid tool calls along the training step. It can be seen that the agent's strategy clearly transitions from \textbf{broad exploration} to \textbf{efficient exploitation}: initially, the agent explores inefficiently, which is reflected in high response lengths and turn counts for a low reward. As training progresses, these trends reverse: the reward consistently rises while the metrics for response length, turns, and invalid tool calls all significantly decrease. This dynamic illustrates a fundamental improvement in the agent's complex question-solving process, not just a superficial optimization of rewards.

\subsubsection{Analysis of Reasoning Capability}

\begin{figure}
    \centering
    \includegraphics[width=1\linewidth]{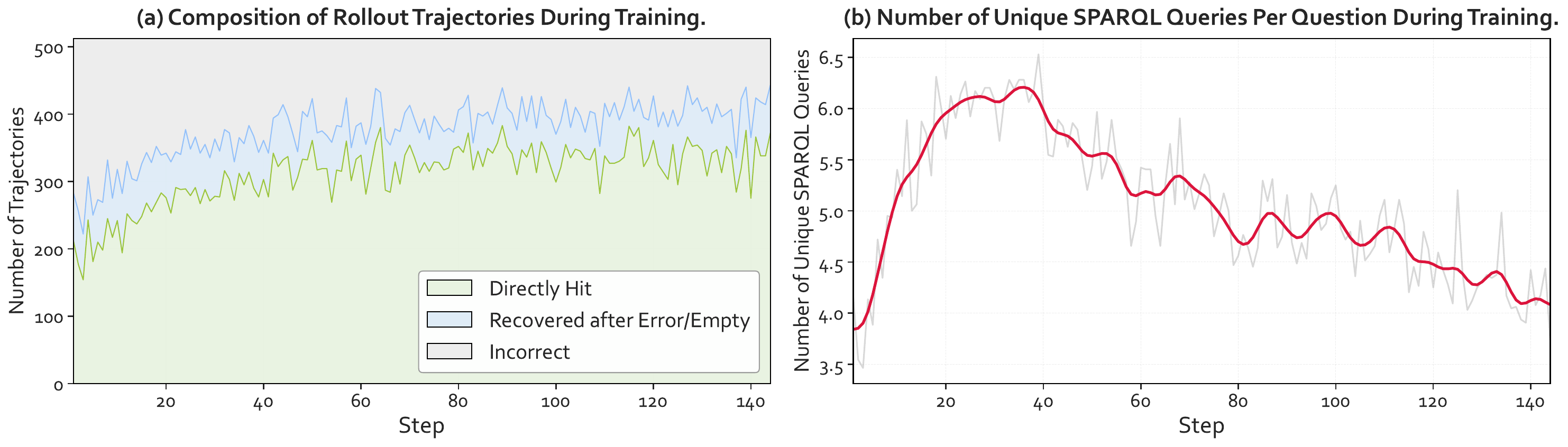}
    \caption{Evolution of Robustness and Flexibility during training: (a) Robustness, shown by the composition of rollout trajectories, and (b) Flexibility, shown by the number of unique SPARQL queries per question.}
    \label{fig:flex_and_rob}
\end{figure}
To provide a deeper analysis of whether the model's reasoning ability is enhanced, particularly concerning the robustness and flexibility lacking in SFT methods, we examine the evolution of the agent's reasoning behavior. Fig.~\ref{fig:flex_and_rob}(a) illustrates its robustness by showing the composition of rollout trajectories, categorized as: "Directly Hit," for trajectories where the final prediction hits at least one gold answers; "Recovered after Error/Empty," for those that hit at least one gold answers after encountering an intermediate error or empty observation; and "Incorrect," for trajectories where the final prediction fails to find any gold answerss. The figure shows that the proportion of successful trajectories ("Directly Hit" and "Recovered") steadily increases during training, This confirms that the agent learns to effectively recover from imperfect feedback, a direct advantage of our outcome supervision approach. In contrast, process supervision encourages the model to simply replicate pre-defined trajectories from the training data, and thus fails to incentivize this type of recovery behavior.

Fig.~\ref{fig:flex_and_rob}(b) measures flexibility by plotting the diversity of trajectories, which we define as the number of unique final SPARQL queries generated for the same question across the rollouts. It can be seen that the agent initially broadens its search for a wide range of solutions before consolidating on a compact set of the most reliable ones, which is consistent with the process discussed in Sec.~\ref{sec:trian_dym}. This indicates that outcome-only supervision naturally fosters an exploration-then-convergence behavior, where the agent ultimately retains sufficient solution diversity to effectively address the lack of flexibility inherent in process-supervised methods.

\subsubsection{Analysis of Process Reward}
To explore whether incorporating process supervision in addition to outcome-only supervision benefits model performance, we experimented by incorporating necessary process-level penalties for behaviors including hallucinations and timeouts. Specifically, we define a hallucination as a tool call containing a relation or type that is not present in the observation history, while a timeout query denotes a call that results in a 300 second timeout. We applied a negative reward of -0.2 for each occurrence of these behaviors, with the total accumulated penalty capped at -0.5. We then compared our full model against two variants incorporating these Process Rewards (P.R.): one applying the penalties throughout the entire RL training (P.R.@RL-phase-1\&2), and another applying them only during the second stage (P.R.@RL-phase-2).
The results clearly demonstrate that introducing penalties for any intermediate behaviors, even for intuitively negative ones, degrades final performance. We hypothesize  that in KBQA tasks, where the correctness of answers can be explicitly verified, process-level rewards may interfere with the agent’s capacity to form a robust trial-and-error exploration strategy, as they tend to prematurely penalize exploratory trajectories that could eventually achieve correct results.
\begin{table}[htbp]
\centering
\caption{Performance comparison of different models and training stages on the WebQSP, CWQ, and GrailQA.}
\label{tab:process}
\begin{tabular}{l cc cc cc}
\toprule
\multirow{2}{*}{\textbf{Model}} & \multicolumn{2}{c}{\textbf{WebQSP}} & \multicolumn{2}{c}{\textbf{CWQ}} & \multicolumn{2}{c}{\textbf{GrailQA}} \\
\cmidrule(r){2-3} \cmidrule(r){4-5} \cmidrule(l){6-7}
 & F1 & Hit@1 & F1 & EM & F1 & EM \\
\midrule
P.R.@RL-phase-1\&2  & 71.5 & 75.3 & 64.9 & 72.7 & 73.0 & 76.2 \\
P.R.@RL-phase-2 & 73.3 & 78.2 & 68.0 & 74.8 & 78.1 & 81.0 \\
\ours         & \textbf{77.2} & \textbf{80.1} & \textbf{68.3} & \textbf{75.7} & \textbf{80.5} & \textbf{83.7} \\
\bottomrule
\end{tabular}
\end{table}

\subsection{Case Study - RQ4}

To concretely demonstrate the advanced reasoning capabilities of \ours, we present a qualitative case study with two detailed examples in Appendix~\ref{app:case}. These examples highlight the agent's ability to: (1) robustly recover from erroneous feedback; (2) flexibly use tools to find correct answers via reasoning trajectories that differ from the gold SPARQL. These cases serve as strong evidence of the robustness and flexibility fostered by the proposed \ours\space approach.

In addition, we conduct error analysis, which is presented in Appendix~\ref{app:error}.

\section{Conclusion}\label{sec:conclusion} In this paper, we focused on a key limitation of existing agentic KBQA paradigms: their reliance on process supervision, which provides weak incentives for autonomous exploration. To overcome this, we proposed \ours, our first agentic reasoning model for KBQA. It is trained by a multi-stage curriculum reinforcement learning framework that effectively leverages outcome-only supervision through a carefully designed learning curriculum. Specifically, \ours\space first establishes foundational reasoning capabilities via a cold-start fine-tuning stage on a small, high-quality dataset created through outcome-based rejection sampling. It then enhances the agent’s exploratory capacity through curriculum RL, employing a progressive reward schedule that transitions from easy to hard tasks. Extensive experiments demonstrate the effectiveness of our approach. Further analyses reveal that \ours\space learns to act as a robust and flexible agent, capable of recovering from errors and strategically exploring diverse reasoning trajectories. Future work may investigate more advanced reflection mechanisms to mitigate remaining error types and extend this curriculum strategy to other complex, agent-based reasoning tasks.

\bibliography{colm2025_conference}
\bibliographystyle{colm2025_conference}

\section{Appendices}

\appendix

\section{Experiment Setup}
\label{app:setup}
\subsection{Datasets}
We evaluate the reasoning performance on three widely-used datasets: WebQSP~\citep{talmor2018web}, CWQ~\citep{talmor2018web}, and GrailQA~\citep{gu2021beyond}. These datasets are constructed based on Freebase and consist of natural language questions paired with their corresponding SPARQL queries. Specifically, WebQSP features questions where the topic entity and answer entities are at most 2 relation hops apart. CWQ extends WebQSP by incorporating more complex entities and constraints, with relation paths up to 4 hops. In contrast to WebQSP and CWQ, GrailQA is designed for a more comprehensive evaluation of generalization capabilities across three settings: independently and identically distributed (I.I.D.), compositional, and zero-shot.

To ensure consistent results and reduce evaluation overhead, we follow the setting of previous work~\citet{xiong2024interactive}, which uniformly sample questions across different types in WebQSP and CWQ. For the GrailQA dataset, we randomly sample 600 instances for evaluation.
\subsection{Baselines}
Our experiments compare \ours \space against two categories of methods.

\textbf{Fine-tuning-based Methods.} We consider two settings: fine-tuning on partial data (low-source setting) and on full data. In the low-source setting, we select the agentic approaches, includes the SFT model in Interactive-KBQA~\citep{xiong2024interactive}, which uses the same toolset as our method (abbreviated as SFT-traj.), two MCTS-based agentic methods, KBQA-o1~\citep{luokbqao1} and MCTS-KBQA~\citep{xiong2025mcts}. Notably, since our method is trained on only a subset of the data and uses no SPARQL annotations, its performance is directly comparable to this category of baselines.
For the full-data setting, which serves as a secondary point of reference, we include the SOTA SP-based methods RnG-KBQA~\citep{ye2022rng} and TIARA~\citep{shu2022tiara}.

\textbf{Prompting-based Methods.} We selected a diverse set of prompting strategies for comparison. These include methods based on direct prompting, Chain-of-Thought (CoT), and self-consistency with CoT; approaches that use an LLM to enhance semantic parsing (KB-BINDER\citep{li2023few} and KB-Coder~\citep{nie2024code}) and subgraph retrieval~\citep{tian2024augmenting}; and Interactive-KBQA~\citep{xiong2024interactive} that prompts an LLM to reason as an agent. For a fair comparison, we report the scores for these methods directly from their official papers, as the specific API versions used in their experiments are no longer accessible.

\subsection{Evaluation Metrics}
We primarily report the F1 score since the answers are returned as an unordered list. Following previous work~\citep{xiong2024interactive}, we also report Random Hits@1 (RHits@1) and Exact Match (EM) for reference. As our method is not fully based on semantic parsing, we do not report the semantically equivalent of logical forms (another score commonly abbreviated as EM) in ~\citet{gu2021beyond}.

\subsection{Implementation Details}
We use the Qwen2.5-Coder-7B-Base as our backbone LLM. For the cold-start stage, our implementation is based on the LlamaFactory framework~\citep{zheng2024llamafactory}; we use a cosine learning rate schedule with an initial learning rate of 2e-5, a batch size of 8, and a maximum sequence length of 4096. We fine-tune the model for 8 epochs. In the RL stage, our implementation is based on Agent-R1 framework \citep{Agent-R1} (a verl-based \citep{sheng2024hybridflow} agentic training framework); we set the number of rollouts to N=8, use a batch size of 64 for rollout, and a global batch size 512 for training. We train the model for 1 epoch. At validation, we set the temperature to 0 to ensure deterministic outputs. The detail parameters can be found in \ref{tab:base_settings}. All experiments were conducted on 8 $\times$ NVIDIA H100 (80GB) GPUs and 8 $\times$ NVIDIA A100 (80GB) GPUs.

\begin{table}[htbp]
\centering
\caption{Key parameters during the training process.}
\label{tab:base_settings}
\begin{tabular}{lc}
\toprule
\textbf{Parameter} & \textbf{Value} \\
\midrule
SFT settings \\
\midrule
Batch size  & 8 \\
Learning Rate & 2e-5 \\
Max Response Length & 4096 \\
Algorithm & LoRA \\
Epoch & 8 \\
\midrule
RL settings \\
\midrule
Rollout Batch Size & 64 \\
Rollout N & 8 \\
Global-batch Size & 512 \\
Learning Rate & 1e-6 \\
Learning Rate Warmup Steps & -1(disabled) \\
Max Prompt Length & 4096 \\
Max Response Length & 8192 \\
KL Loss Coefficient & 0.001 \\
Algorithm & GRPO \\
Epoch & 1\\
Validation Temperature & 0.0 \\
Rollout Temperature & 1.0 \\
\bottomrule
\end{tabular}
\end{table}
\section{Further Analysis}
In this section, we provide several supplementary analyses. First, we examine the impact of different backbone models on performance and discuss our method's token efficiency. These results justify our model selection and highlight the efficiency advantages of our approach. Second, we present a joint distribution of error frequency versus sample accuracy across training stages, which serves to supplement our main analysis of training dynamics (Section~\ref{sec:trian_dym}) and offers a more comprehensive view of how the agent's capabilities evolve.
\subsection{Analysis of the Impact of Different Backbones}
We compared the impact of different backbone models on performance during the cold-start stage. As shown in Table~\ref{tab:model_performance}, our results indicate that Qwen2.5-Coder-7B-Base delivers the strongest overall performance. Although Qwen3-8B-Base performs better on the simpler WebQSP dataset, we selected Qwen2.5-Coder-7-Base for its superior performance on CWQ, which contains more complex questions. Furthermore, our task involves numerous code-style tool calls, where a coder-specialized model has a natural advantage. 
\begin{table}
  \centering
  \caption{Performance comparison of various backbone models across three benchmark datasets.}
  \label{tab:model_performance}
  \begin{tabular}{lcccccc}
    \toprule
    \multirow{2}{*}{\textbf{Model}} & \multicolumn{2}{c}{\textbf{WebQSP}} & \multicolumn{2}{c}{\textbf{CWQ}} & \multicolumn{2}{c}{\textbf{GrailQA}} \\
    \cmidrule(lr){2-3} \cmidrule(lr){4-5} \cmidrule(lr){6-7}
    & F1   & Hits@1 & F1   & EM   & F1   & EM   \\
    \midrule
    Qwen2.5-Coder-3B-Base & 58.4 & 63.9 & 36.7 & 41.2 & 58.4 & 68.2 \\
    Qwen2.5-Coder-7B-Base & 56.7 & 64.9 & 50.5 & 59.8 & 63.1 & 68.4 \\
    Qwen3-4B-Base    & 63.9 & 69.4 & 38.4 & 39.0 & 55.0 & 59.1 \\
    Qwen3-8B-Base    & 65.3 & 71.6 & 44.6 & 45.8 & 60.1 & 63.1 \\
    \bottomrule
  \end{tabular}
\end{table}
\subsection{Analysis of token efficiency}
To evaluate the inference efficiency of our method, we conducted a computational cost analysis against prompting-based baselines, including the single-round inference method CoT and the multi-round interactive method ToG. As shown in Table~\ref{tab:cost_hit}, the comparison covers LLM call frequency, token usage, and overall cost. The results reveal that \ours \space achieves performance comparable to the complex ToG, while maintaining a computational budget similar to the much simpler CoT. This efficiency stems from outcome-based supervision: the model learns to recover from empty results or intermediate errors in a single pass, obviating the need to generate multiple branches to find a solution at significant computational expense. Overall, \ours \space shifts the efficiency frontier toward lower cost while maintaining strong effectiveness, making it more predictable and easier to deploy in real-world scenarios.
\begin{table}[ht]
  \centering
  \small
  \caption{Cost and effectiveness on CWQ and WebQSP.}
  \label{tab:cost_hit}

  \begin{tabular}{lcccc}
    \toprule
    \multicolumn{5}{c}{\textit{CWQ}} \\
    \midrule
    Model & \# Avg. LLM Call & Total Token & Total Cost & Hit@1 \\
    \midrule
    CoT  & 1.0 & 409.7   & \(8.00\times10^{-5}\) & 43.7 \\
    ToG  & 9.2 & 11468.5 & \(2.30\times10^{-3}\) & 69.5 \\
    Ours & \underline{5.0} & \underline{660.6} & \textbf{\(0\)} & \textbf{75.7} \\
    \bottomrule
  \end{tabular}

  \begin{tabular}{lcccc}
    \toprule
    \multicolumn{5}{c}{\textit{WebQSP}} \\
    \midrule
    Model & \# Avg. LLM Call & Total Token & Total Cost & Hit@1 \\
    \midrule
    CoT  & 1.0 & 397.6   & \(8.00\times10^{-5}\) & 47.5 \\
    ToG  & 8.8 & 10189.4 & \(2.10\times10^{-3}\) & 82.6 \\
    Ours & \underline{3.9} & \underline{439.6} & \textbf{\(0\)} & \underline{80.1} \\
    \bottomrule
  \end{tabular}
\end{table}
\subsection{Analysis on Overall Robustness Capability Shift}
\begin{figure}[!t]
    \centering
    \includegraphics[width=0.9\linewidth]{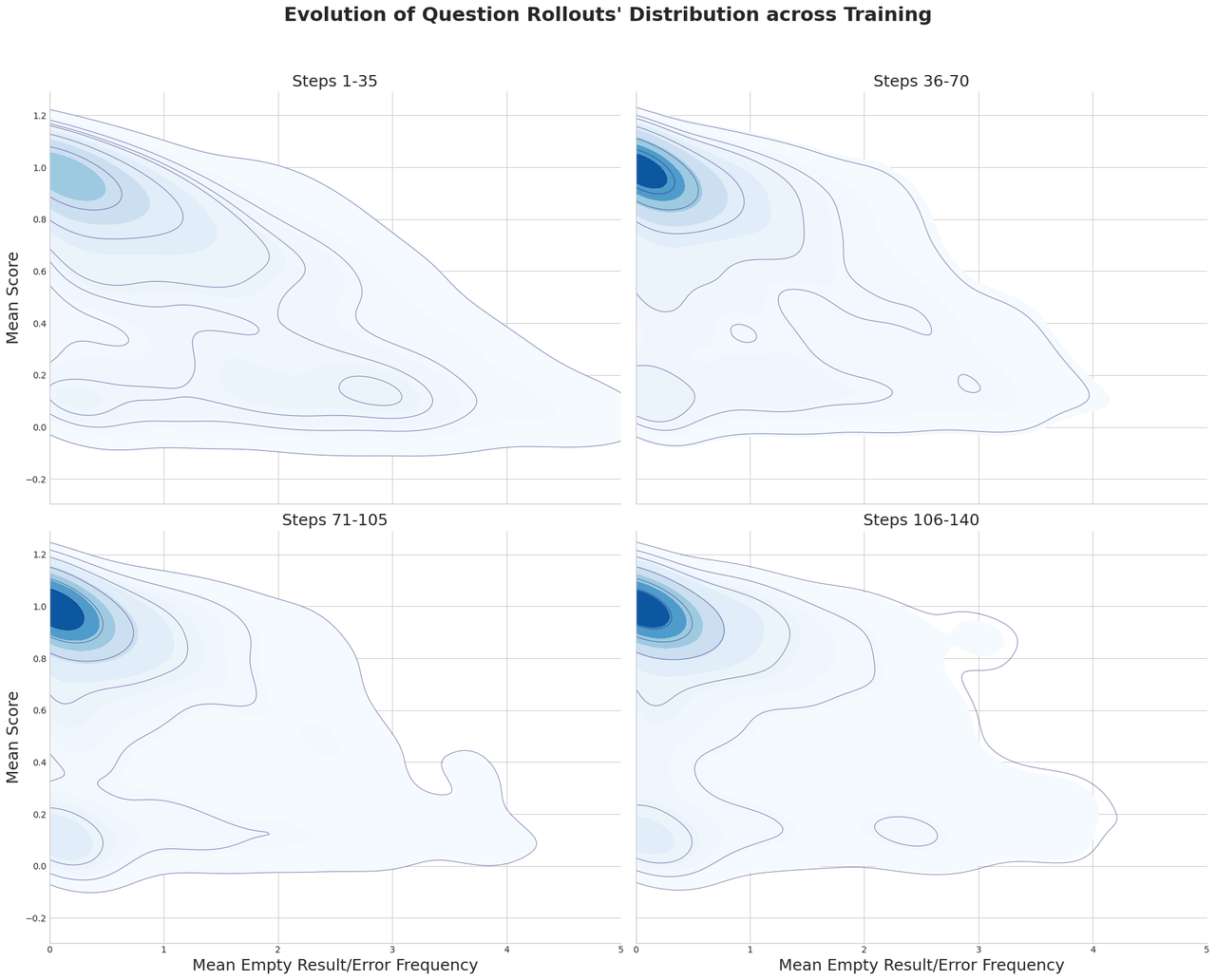}
    \caption{Joint distribution of mean empty-result/error frequency and mean reward across training phases.}
    \label{fig:union_dynamic}
\end{figure}

To further investigate how the agent's robustness evolves, we analyze the association between intermediate errors and final rewards throughout training. As illustrated in Figure~\ref{fig:union_dynamic}, we plot the joint distribution of the mean frequency of empty results/errors per question against the mean reward. This visualization reveals a clear progression. Early in training, the distribution forms a downward-sloping triangle, where higher error rates are strongly correlated with lower rewards. As training advances, this shape flattens into a rectangle, indicating that high rewards are achieved even with a non-trivial frequency of errors. Concurrently, the density increases in the upper-left corner, reflecting a growing number of questions being solved directly and perfectly. The transformation provides compelling evidence of our method's superiority: the agent not only learns to avoid mistakes but also develops the crucial ability to recover from them, transitioning from a brittle execution policy to a robust, high-scoring one.

\section{Statistics}
\label{app:stat}
We give the statistic of training samples used in cold-start-stage and RL stage on three datasets. 
\begin{table}[h]
\centering
\caption{Number of samples for different training stages across the WebQSP, CWQ, and GrailQA datasets.}
\label{tab:training_samples}
\begin{tabular}{l ccc}
\toprule
\textbf{Stage} & \textbf{WebQSP} & \textbf{CWQ} & \textbf{GrailQA} \\
\midrule
RL-Phase-1  & 1265 & 2400 & 1800 \\
RL-Phase-2  & 709  & 1600 & 1200 \\
Cold-Start & 440  & 699  & 563  \\
\bottomrule
\end{tabular}
\end{table}

\section{Error Analysis}
\label{app:error}
\begin{figure*}[!t]
    \centering
    \includegraphics[width=1\linewidth]{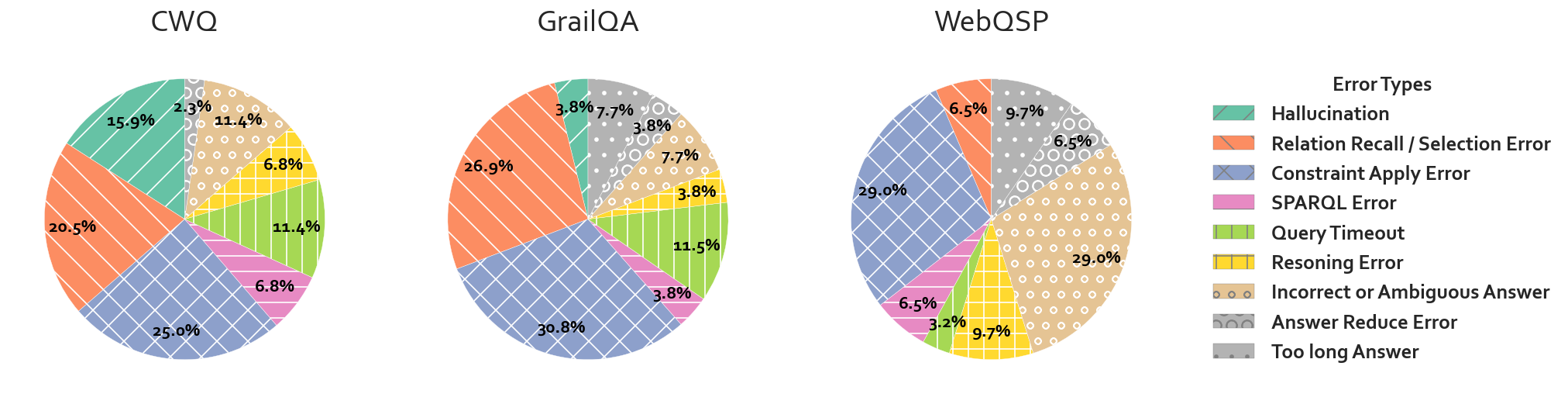}
    \caption{Analysis of sampled failure cases per dataset. We visualize the proportion of factors contributing to errors.}
    \label{fig:error}
\end{figure*}
To identify the limitations of our method, we conducted an error analysis by manually inspecting random sampled cases where the $F_1$ score was less than 1. The results are categorized in Fig.~\ref{fig:error}, with the primary error types being Hallucination, Constraint Application Error, and Relation Recall/Selection Error. Hallucinations often result from flawed self-correction attempts. While intuitively detrimental, our prior experiments show that penalizing such behavior actually stifles the model's capacity for self-correction. The other two error types generally stem from incorrect decisions during complex reasoning. These errors highlight the space for future work: enhancing the model's ability to handle complex questions within a linear decision-making process. Notably, the high rate of constraint errors on WebQSP is due to the model failing to identify implicit temporal constraints (e.g., the KB's 2015 cutoff date) in present-tense questions.
\section{Prompts}
\label{app:prompt}

In our experiments, the same prompt was used for both training and evaluation, which is composed of a system prompt and a user prompt.
\begin{examplebox}{System Prompt}
\begin{lstlisting}
#Tools

You are an expert in knowledge base query language SPARQL programming. The user gives a question, and you need to iteratively call the tool to continuously improve the SPARQL query until it can get the answer to the question.

You are provided with function signatures within <tools></tools> XML tags:
<tools>
{'type': 'function', 'function': {'name': 'SearchGraphPatterns', 'description': 'This tool searches for relevant one-hop and two-hop subgraphs tied to a specified variable. It queries subgraphs where the chosen variable (?x, assuming the SPARQL query begins with "SELECT DISTINCT ?x WHERE") appears as the head or tail entity and returns them collectively. The semantic parameter indicates the expected predicate semantics. When provided, the tool ranks the subgraphs based on these semantics. If unspecified, it returns the complete subgraph.', 'parameters': {'type': 'object', 'properties': {'sparql': {'type': 'string', 'description': 'SPARQL query'}, 'semantic': {'type': 'string', 'description': 'The semantic parameter represents the expected predicate semantics.'}}, 'required': ['sparql']}}}

{'type': 'function', 'function': {'name': 'ExecuteSPARQL', 'description': 'This tool executes a SPARQL query and returns the results.', 'parameters': {'type': 'object', 'properties': {'sparql': {'type': 'string', 'description': 'SPARQL query'}}, 'required': ['sparql']}}}

{'type': 'function', 'function': {'name': 'SearchTypes', 'description': 'Search the knowledge base for matching semantic types, used to initiate queries from a type when no topic entities are available, or to find a type to refine the query when multiple entities are returned. When use the type, please give the sparql as: SELECT DISTINCT ?x WHERE { ?x ns:type.object.type ns:<type_name> }', 'parameters': {'type': 'object', 'properties': {'query': {'type': 'string', 'description': 'the semantic of type to search for'}}, 'required': ['query']}}}
</tools>

For each function call, return a json object with function name and arguments within <tool_call></tool_call> XML tags:
<tool_call>
{{"name": <function-name>, "arguments": <args-json-object>}}
</tool_call>
\end{lstlisting}
\end{examplebox}

\begin{examplebox}{User Prompt}
\begin{lstlisting}
When you encounter a complex question, you should break it down into several sub-questions and answer them step by step. You can use the tools provided. You can use the tool as many times as you want.
You must first conduct reasoning inside <think>...</think>. If you need to use the tool, you can use the tool call <tool_call>...</tool_call> to call the tool after <think>...</think>.
When you have the final answer, you can output the answer in the python list format inside <answer> tag, such as: <answer> the answer is \\boxed{{[...]}} </answer>.

Output format for tool call:
<think>
...
</think>
<tool_call>
...
</tool_call>

Output format for answer:
<think>
...
</think>
<answer>
...
</answer>

Question: {question}
Topic Entities: {topic entities}
Assistant:
\end{lstlisting}
\end{examplebox}
\section{Case Study}
\label{app:case}
To illustrate the flexibility and robustness of our agent's reasoning process, we present two representative case studies. The first case demonstrates robustness: when faced with two consecutive empty query results, the agent successfully recovers by leveraging a more distant observation from its history to try a new, promising relation, ultimately finding the correct answer. The second case highlights flexibility and self-correction, showing the agent's ability to identify an illogical query result (e.g., receiving a high school when a university was expected) and adapt its strategy by employing a novel action to resolve the issue.

\begin{table*}
  \caption{Example of robust reasoning of \ours.}
  \label{tab:case-study-1}
  \begin{center}
  \begin{tabularx}{\textwidth}{ L }
  \toprule
  \textbf{Question:} who plays london tipton in suite life on deck? \\
  \midrule
  \thinkrow{I need to find the actor who plays London Tipton in "The Suite Life on Deck". I will start by searching for predicates related to "actor" or "performance" for the entity "London Tipton(m.07g8r3)".}
  \searchrow{SearchGraphPatterns(sparql="SELECT ?e WHERE \{ VALUES ?e \{ns:m.07g8r3\} \}", semantic="actor/performer")}
  \resultrow{[..., (?e, film.film\_character.portrayed\_in\_films -> film.performance.actor, "Brenda Song"),...]}
  \thinkrow{(Fail Try 1)...}
  \searchrow{ExecuteSPARQL(sparql="SELECT ?actor WHERE \{ VALUES ?e1 \{ns:m.07g8r3\} . VALUES ?e2 \{ns:m.03mj4jm\} . ?e1 ns:film.film\_character.portrayed\_in\_films ?cvt . ?cvt ns:film.performance.actor ?actor . ?cvt ns:film.performance.film ?e2 . \}")}
  \resultrow{[]}
  \thinkrow{\textbf{The previous query returned no results. This might be due to the structure of the predicates or the way the entities are linked. I will try a different approach by searching for predicates related to "actor" or "performance"}...}
  \searchrow{SearchGraphPatterns(sparql="SELECT ?e WHERE \{ VALUES ?e \{ns:m.03mj4jm\} \}", semantic="actor/performer/character")}
  \resultrow{[...,(?e, award.award\_nominated\_work.award\_nominations -> award.award\_nomination.award, "Kids' Choice Award for Favorite TV Actor"), ...]}
  \thinkrow{(Fail Try 2)...}
  \searchrow{ExecuteSPARQL(sparql="SELECT ?actor WHERE \{ VALUES ?e1 \{ns:m.07g8r3\} . VALUES ?e2 \{ns:m.03mj4jm\} . ?e2 ns:award.award\_nominated\_work.award\_nominations ?cvt . ?cvt ns:award.award\_nomination.award\_nominee ?actor . ?cvt ns:award.award\_nomination.character ?e1 . \}")}
  \resultrow{[]}
  \thinkrow{\textbf{The previous queries have not yielded the correct result. I need to refine my approach.} The question asks for... I will try a different strategy...I will try to find the actor who played London Tipton in any film and then filter for "The Suite Life on Deck".}
  \searchrow{ExecuteSPARQL(sparql="SELECT ?actor WHERE \{ VALUES ?e \{ns:m.07g8r3\} . ?e ns:film.film\_character.portrayed\_in\_films ?cvt . ?cvt ns:film.performance.actor ?actor . \}")}
  \resultrow{["Brenda Song"]}
  \answerrow{Brenda Song} 
  \bottomrule
  \end{tabularx}
  \end{center}
\end{table*}

\begin{table*}[!t]
  \caption{Example of flexible reasoning of \ours.}
  \label{tab:case-study-2}
  \begin{center}
   \begin{tabularx}{\textwidth}{ L }
  \toprule
  \textbf{Question:} Where did Charles Drew attend \textbf{college} that has the latest founding date? \\
  \midrule
  \thinkrow{I need to find the colleges that Charles R. Drew attended and then find the college with the latest founding date.}
  (Omit for space...)
  \\
  \searchrow{ExecuteSPARQL(sparql="SELECT DISTINCT ?college WHERE \{ VALUES ?e \{ns:m.018t67\} . ?e ns:people.person.education ?edu . ?edu ns:education.education.institution ?college . ?college ns:organization.organization.date\_founded ?founded\_date . \} ORDER BY DESC(xsd:date(?founded\_date)) LIMIT 1")}
  \resultrow{["Dunbar High School"]}
  \thinkrow{\textbf{The result is "Dunbar High School", but the question asks for the college, not high school. I need to re-evaluate the search.} ... I will search for the type "College/University".}
  \searchrow{SearchTypes(query="College/University")}
  \resultrow{["education.university", ...]}
  \thinkrow{The "SearchTypes" results include "education.university" ... I will find the colleges attended by Charles R. Drew and their founding dates, then order by founding date in descending order and limit to 1 to get the college with the latest founding date.}
  \searchrow{ExecuteSPARQL(sparql="SELECT DISTINCT ?college WHERE \{ VALUES ?e \{ns:m.018t67\} . ?e ns:people.person.education ?edu . ?edu ns:education.education.institution ?college . ?college ns:type.object.type ns:education.university . ?college ns:organization.organization.date\_founded ?founded\_date . \} ORDER BY DESC(xsd:date(?founded\_date)) LIMIT 1")}
  \resultrow{["McGill University Faculty of Medicine"]}
  \answerrow{McGill University Faculty of Medicine} 
  \bottomrule
  \end{tabularx}
  \end{center}
\end{table*}


\end{document}